\documentclass{article} 
\usepackage{iclr2015,times}
\usepackage{hyperref}
\usepackage{url}
\usepackage{graphicx}

\title{Gradual Training of Deep Denoising Auto Encoders}

\author{
Alexander Kalmanovich \& Gal Chechik \\ 
The Gonda Brain Research Center\\
Bar Ilan University\\
52900 Ramat-Gan, Israel\\
\texttt{sashakal@gmail.com, gal.chechik@biu.ac.il} \\
}

\newcommand{\first}{1\textsuperscript{st}}
\newcommand{\second}{2\textsuperscript{nd}}
\newcommand{\third}{3\textsuperscript{rd}}

\newcommand{\ignore}[1]{}

\iclrconference 

\begin{document}

\maketitle

\begin{abstract}
Stacked denoising auto encoders (DAEs) are well known  to learn useful deep representations, which can be used to improve supervised training by initializing a deep network. We investigate a training scheme of a deep DAE, where DAE layers are gradually added and keep adapting as additional layers are added. We show that in the regime of  mid-sized datasets, this \textbf{\textit{gradual training}} provides a small but consistent improvement over stacked training in both reconstruction quality and classification error over stacked training on MNIST and CIFAR datasets.
\end{abstract}

\section{Introduction}
A central approach in learning meaningful representations is to train a deep network for reconstructing corrupted data. The idea is simple: given unlabeled data, a deep-network is given input-output pairs, where the input consists of a corrupted version of an input sample and the output consists of the original non-corrupted version which the network aims to reconstruct. Indeed, \textbf{\textit{denoising autoencoders}} (DAE)~\citep{vincent2008extracting} have been shown to extract meaningful features which allow to correct corrupted input data~\citep{xie2012image}. These representations can later be used to initialize a deep network for a supervised learning task. It has been shown that in the small-data regime, good initializations can cut down the training time and improve the classification accuracy of the supervised task~\citep{vincent2010stacked,vincent2008extracting,larochelle2009deep,erhan2010does}.

Going beyond a single layer, it has been shown that training a multi-layer (deep) DAE can be achieved efficiently by stacking single-layer DAEs and training them layer-by-layer~\citep{vincent2010stacked}. Specifically, a stacked denoising autoencoder (SDAE) is trained as follows (Fig.~\ref{figure1}). First, a single-layer auto encoder is trained over the corrupted input data $\tilde{x}$ and its weights  are tuned (Fig.~\ref{figure1}a). Then, the weights to the first hidden layer $w_1$ are frozen, and the data is transformed to the hidden representation (Fig.~\ref{figure1}b). This transformed input $h_1(x)$ is then used to create a corrupted input to a second autoencoder and so on (Fig.~\ref{figure1}c). 

Stacked training has been shown to outperform training de-novo of a full deep network, presumably because it provides better error signals to lower layers of the network~\citep{erhan2009difficulty}. However, stacked training is greedy in the following sense: When the first layer is trained, it is tuned such that its features can be directly used for reconstructing the corrupted input. Later on however, these features are used as input to train more complex features. Comparing this with the process of reduced plasticity in natural neural systems, early layers in mammalian visual system keep adapt for prolonged periods, and their synapses remain plastic long after representations have been formed in high brain areas~\citep{liu2004switching}. We therefore turned to explore alternative training schedules for deep DAEs, which avoid freezing early weights.

We test here \textbf{\textit{‘gradual training’}}, where training occurs layer-by-layer, but lower layers keep adapt throughout training. We compare gradual training to stacked training and to a hybrid approach, all under a fixed budget of training update steps. We then test gradual training as an initialization for supervised learning, and quantify its performance as a function of dataset size. Gradual training provides a small but consistent improvement in reconstruction error and classification error in the regime of mid-sized datasets.

\section{Training denoising autoencoders}
\begin{figure}[h]
    \centering
    \includegraphics[clip, trim = 0cm 6cm 0cm 5cm, height=5.5cm]{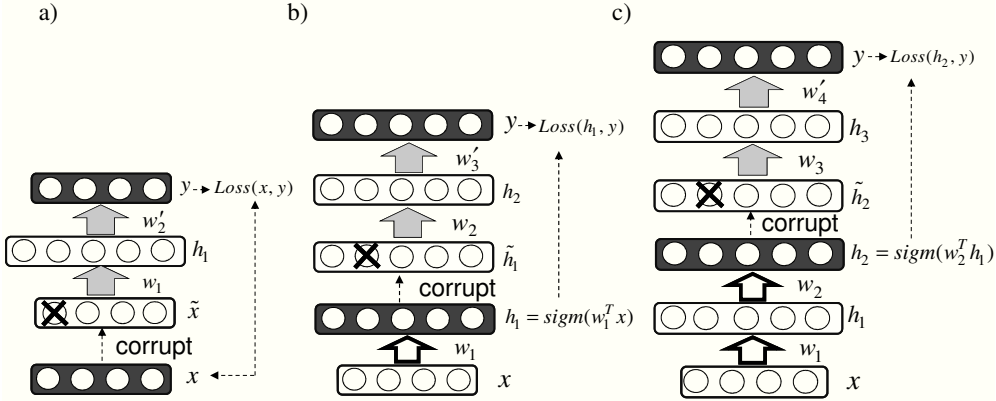}
    \caption{Stacked-training of a stacked DAE with 3 hidden layers. $x$ denotes an input sample and $y$ denotes the network output. Black layers are the ones used for computing the loss. Gray arrows denote weights that are updated through back propagation, while blank arrows denote denote weights which are not changed during training. $w'$ weights are discarded in subsequent training phases. Crosses illustrate corrupted units.  \textbf{(a)} Training the \first{} hidden layer. \textbf{(b)} Training the \second{} hidden layer. Noise is injected to $h_1$, creating  $\tilde{h_1}$. \textbf{(c)} Training the \third{} hidden layer. }
    \label{figure1}
\end{figure}

For completeness, we detail here the procedure for training stacked denoising autoencoders described by \citet{vincent2010stacked}. Fig.~\ref{figure1} describes the architecture and the main training phases. For training the first layer with a training sample $x$, masking noise is used to create a corrupted noisy version $\tilde{x}$ (Fig.~\ref{figure1}a, “corrupt” arrow). A forward pass is taken, computing the hidden representation $h_1=Sigmoid(w^\top_1x)$ and the output $y=Sigmoid({w_2'^\top}h_1)$. Specifically, the loss function is often taken to be the cross entropy between $y$ and $x$ (Fig.~\ref{figure1}a, dotted arrow). All weights are updated by propagating the error gradient back through the network. This is repeated for other samples in a stochastic gradient descent (SGD) fashion, and combined with momentum and weight decay to speed training. 

To train a deep network, multiple DAEs are stacked using greedy layer-wise training~\citep{vincent2010stacked}. After the first DAE is trained, the learned encoding weights $w_1$ are fixed, and the data is mapped to the hidden layer representation $h_1$ (Fig.~\ref{figure1}b, blank arrow). The second DAE is trained based on $h_1(x)$ using the same procedure as the first layer (Fig.~\ref{figure1}b). Importantly, the corrupting noise is applied to the hidden representation $h_1(x)$ to create $\tilde{h_1}$, with the motivation being that injecting noise to the hidden layer introduces variability of the more-abstract representation that was already learned by the network. Training of subsequent layers follows the same procedure, injecting noise at higher and higher layers. 

Often, this layer-wise training procedure is followed by a full back-propagation phase, where noise is injected to the original input $x$ and all layers are updated jointly. Then, the SDAE can be used to initialize a deep network for a supervised classification task by replacing the top reconstruction layer with a (usually multi-class) classification layer. 

\subsection{Gradual training of deep DAEs}
\begin{figure}[h]
    \centering
    \includegraphics[clip, trim = 0cm 6cm 0cm 5.05cm, height=5.25cm]{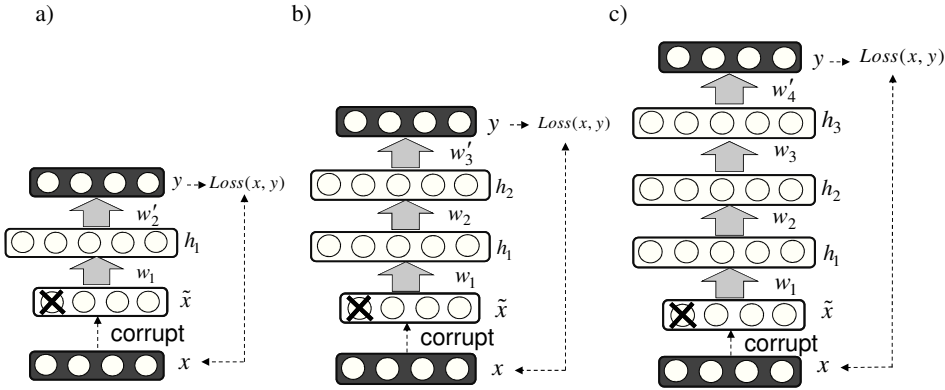}
    \caption{Gradual training of denoising auto encoder with 3 hidden layers. \textbf{(a)} Training 1\textsuperscript{st} hidden layer. \textbf{(b)} Training layers 1 + 2. \textbf{(c)}  Training layers 1 + 2 + 3. In all panels, $x$ denotes an input sample and $y$ the network output. The loss is computed over the black layers. Gray arrows denote weights that are updated through back propagation. $w'$ denotes weights used for decoding, and are discarded in subsequent training phases. Crosses illustrate corrupted units.}
    \label{figure2}
\end{figure}

We describe an alternative, {\em{\textbf{gradual}}}, scheme for training autoencoders. The basic idea is to train the deep autoencoder layer-by-layer, but keep adapting the lower layers continuously. Noise injection is only applied at the input level (Fig.~\ref{figure2}). The motivation for this procedure has two aspects. First, it allows lower weights to take into account the higher representations during training, reducing the greedy nature of stacked training. Second, denoising is applied to the input, rather than to a hidden representation learned in a greedy way. 

More specifically, the first layer is trained in the same way as in stacked training, producing the weights $w_1$. Then, when adding the second layer autoencoder, its weights $w_2$ are tuned jointly with $w_1$. This is done by using the weights $w_1$ to initialize the first layer and randomly initializing the weights of the second. Given a training sample $x$, we generate a noisy version $\tilde{x}$, feed it to the 2-layered DAE, and compute the activation at the subsequent layers
$h_1=Sigmoid(w^\top_1x)$, $h_2=Sigmoid(w^\top_2h_1)$ and
$y=Sigmoid({w_3'^\top}h_2)$. Importantly, the loss function is now computed over the input $x$, and is used to update all the weights including $w_1$ (Fig.~\ref{figure2}b). Similarly, if a \third{} layer is trained, it involves tuning $w_1$ and $w_2$ in addition to $w_3$ and $w'_4$ (Fig.~\ref{figure2}c).

There are therefore two main differences between gradual and stacked training of SDAE. First, in gradual training, weights of lower layers are never fixed as in stacked training, but rather trained jointly when tuning weights of a newly-added layer. Second, each training phase reconstructs a noisy version of the input rather than a noisy version of a hidden-layer representation.

\section{Experiments}
We compare the performance of gradual training of DAEs with that of stacked training,  in two learning setups. First in an unsupervised denoising task, and then by using them to initialize a deep network in a supervised classification task.
We conduct all experiments using "MEDAL", a MATLAB implementation of DNNs and auto encoders~\citep{DustinMedal}.

\subsection{Datasets}
We tested gradual training on three benchmark datasets: MNIST~\citep{lecun1998gradient}, CIFAR-10 and CIFAR-100~\citep{krizhevsky2009learning}. MNIST contains 70,000 28-by-28 grayscale images, each containing a single hand-written digit.  CIFAR-10 and CIFAR-100 contain 60,000 natural RGB images of 32-by-32 pixels from 10 or 100 categories respectively.

\subsection{Training procedure}
Performance was evaluated on a test subset of 10,000 samples. When quantifying performance as a function of dataset size, we create training subsets of different sizes while maintaining the class distribution uniform as in the original training data. 


Hyper parameters were selected using a second level of cross validation (10-fold CV for MNIST, 5-fold for CIFAR), keeping a uniform distribution over classes. In the experiments below, we tune the following hyper parameters: number of units in hidden layers (same for all layers: $1000$,$1500$,$2000$,$2500$), learning rate ($10^{-1}$, $10^{-2}$, $10^{-3}$, $5\times10^{-4}$, $10^{-4}$, $5\times10^{-5}$, $10^{-5}$) batch size for SGD ($10,20$), seed for weight random initialization, momentum ($0.9,0.7$)~\citep{polyak1964some} and weight decay ($10^{-3}$, $10^{-4}$, $10^{-5}$)~\citep{moody1995simple}. The best performing configuration on the validation set was sought in a semi-automatic fashion (as in \citet{vincent2010stacked}) by running experiments in parallel on a large computation cluster with manual guidance to avoid wasting resources on unnecessary parts of the configuration space. We used early stopping by monitoring reconstruction error or classification error on the validation set, and stopped training after 35 epochs without improvement. We used the parameters (weights) which yield the best performance over the validation set. Reported results are the average over 3 different random train-validation splits.

Since gradual training involves updating lower layers, every presentation of a sample involves more weight updates than in a single-layered DAE. We compare stacked and gradual training on a common ground, by using the same ‘budget’ for weight update steps. For example, when training the second layer for $n$ epochs in gradual training, we allocate $2n$ training epochs for stacked training. The overall budget for update steps was determined using early stopping, such that the reconstruction error on the validation set in the last 10 epochs did not improve more than 0.5\% in all training schemes.

\section{Results}
We evaluate gradual and stacked training in unsupervised task of image denoising. We then test these training methods as an initialization for supervised learning, and quantify its performance as a function of dataset size.
\subsection{Unsupervised learning for denoising}
We start by evaluating gradual training in an unsupervised task of image denoising. Here, the network is trained to minimize a cross-entropy loss over corrupted images.   
In addition to stacked and gradual training, we also tested a hybrid method that spends some epochs on tuning only the second layer (as in stacked training), and then spends the rest of the training budget on both layers (as in gradual training). We define the \textit{Stacked-vs-Gradual} fraction $0\leq f \leq1$ as the fraction of weight updates that occur during ‘stacked’-type training. $f=1$ is equivalent to pure stacked training while $f=0$ is equivalent to pure gradual training. Given a budget of n training epochs, we train the \second{} hidden layer with gradual training for $n(1-f)$ epochs, and with stacked training for $2nf$ epochs. 

\begin{figure}[h]
    \centering
    \includegraphics[clip, trim = 0cm 10.5cm 0cm 0cm, height=4.92cm]{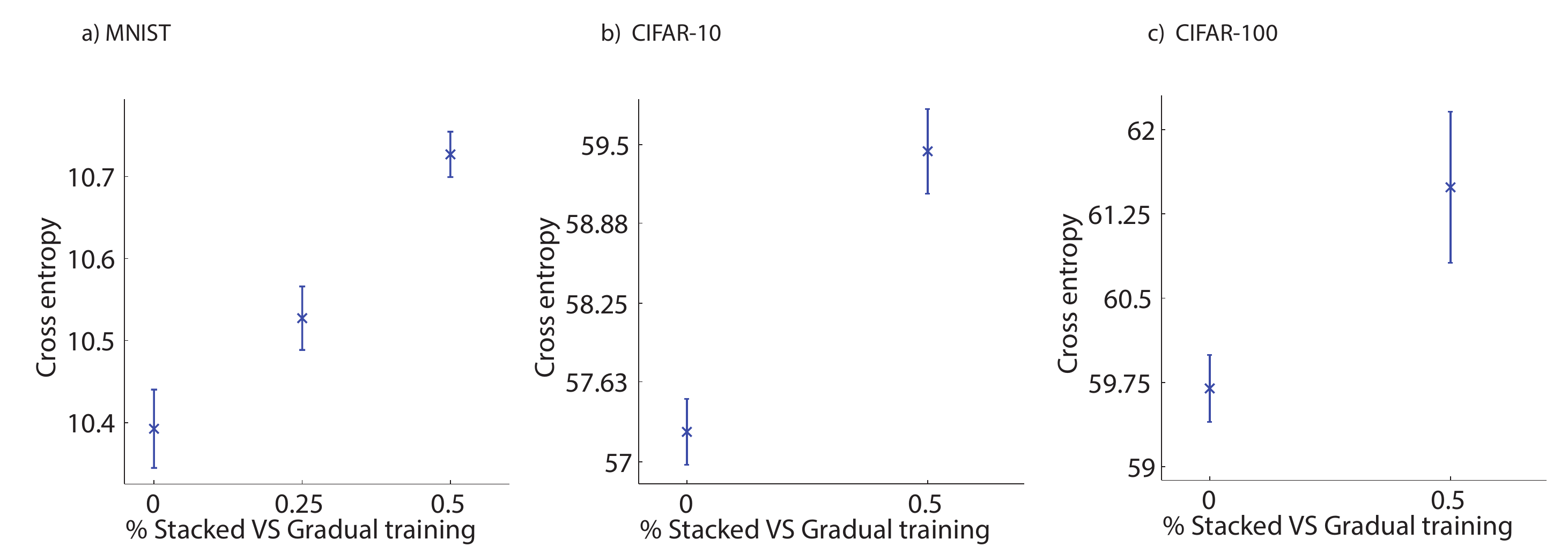}
    \caption{Reconstruction error of unsupervised training methods measured by cross-entropy loss. Error bars are over 3 train-validation splits. The shown cross-entropy error is relative to the minimum possible error, computed as the cross-entropy error of the original uncorrupted test set with itself. All compared methods used the same budget of update operations.
\textbf{(a)} MNIST dataset. Images were corrupted with 15\% masking noise. Network has 2 hidden layers with 1000 units each. The \first{} hidden layer is trained for 50 epochs. Total epoch budget for the \second{} hidden layer is 80 epochs. \textbf{(b)}  CIFAR-10 dataset. Images were corrupted with 10\% masking noise. Network architecture: 2 hidden layers, each with 1500 units. The \first{} hidden layer is trained for 25 epochs. Total epoch budget for \second{} hidden layer is 70 epochs. \textbf{(c)} CIFAR-100 dataset. Noise corruption level is 10\%. Network architecture is 2 hidden layers with 2500 units each. 1st hidden layer is trained for 35 epochs. Total epoch budget for \second{} hidden layer is 70 epochs. }
\label{figure3}
\end{figure}

Figure~\ref{figure3} shows the test-set cross entropy error when training 2-layered DAEs, as a function of the \textit{Stacked-vs-Gradual} fraction. Pure gradual training achieved significant lower reconstruction error than any mix of stacked and gradual training with the same budget of update steps. 

We also evaluated the reconstruction error after a full tuning phase is performed in which all weights are updated jointly for $80$ epochs for MNIST and $70$ epochs for CIFAR. Pure gradual training ($f=0$) improved the reconstruction error over full stacked training  ($f=1$) across all datasets. MNIST: from $10.38\pm0.06$ to $9.61\pm0.06$, being a $7.39\%$ improvement; CIFAR-10: from $57.26\pm0.23$ to $55.47\pm0.2$, being a $3.12\%$ improvement; CIFAR-100: from $59.34\pm0.3$ to $57.01\pm0.45$, being a $3.92\%$ improvement. 

\subsection{Gradual-training DAE for initializing a network in a supervised task}
We use DAEs trained in the previous experiment for initializing a deep network to solve a supervised classification task. The network architecture is the same as SDAE architecture, except for the top layer. The first two hidden layers are initialized with the first two layer weights of the SDAE ($w_1$ and $w_2$ in Fig.~\ref{figure2}b). We then add a top classification layer with output units matching the classes in the dataset, with randomly initialized weights. 

\begin{figure}[h]
\centering
\includegraphics[clip, trim = 0cm 10cm 0cm 0cm, height=5.38cm]{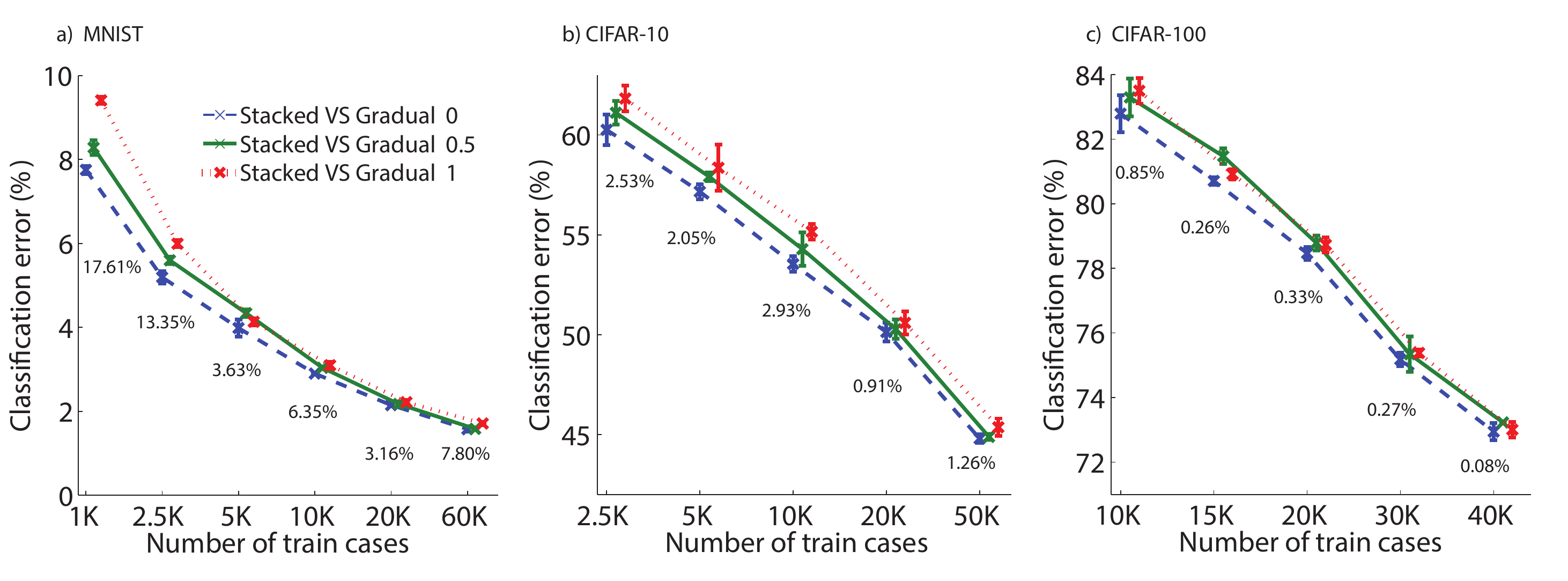}
\caption{Classification error of supervised training initialized based on DAEs. Error bars are over 3 train-validation splits. Each curve shows a different pre-training type (see Fig.~\ref{figure2}). Text labels show the percentage of error improvement of \textit{Stacked-vs-Gradual 0} pretraining (Fig.~\ref{figure2}) compared to \textit{Stacked-vs-Gradual 1} pretraining (not shown in Fig.~\ref{figure2}). \textbf{(a)} MNIST. Two hidden layers with 1000 units each. \textbf{(b)} CIFAR-10. Two hidden layers with 1500 units each. \textbf{(c)} CIFAR-100. Two hidden layers with 2500 units each.}
\label{figure4}
\end{figure}

We train these networks on several subsets of each dataset to quantify the benefit of unsupervised pretraining as a function of train-set size. Figure.~\ref{figure4} traces the classification error as a function of training set size, showing in text the percentage of relative improvement. These results  suggest that initialization with gradually-trained DAEs yields better classification accuracy than when initializing with stacked-trained DAEs, and that this effect is mostly relevant for datasets with less than $50K$ samples. 

The gradual training procedure described above differs from stacked training in two aspects: noise injection at the input level and joint training of weights. To test which of these two contributes to the superior performance we conducted the following experiment. We trained a network to reconstruct a noisy version of the input, as in gradual training, but kept the weights of the 1\textsuperscript{st}  hidden layer fixed as in stacked training. 

The results of this experiments varied across datasets. In MNIST, injecting noise to the input while freezing the first layer performed worse than gradual training, both in terms of cross entropy (in the reconstruction task) and in terms of classification accuracy (in the supervised task). In CIFAR however, training with freezing the first layer actually reduced reconstruction error compared with gradual training, while achieving the same performance in the supervised task.

\section{Conclusion}
We described a ‘gradual training’ scheme for denoising auto encoders, which improves the reconstruction error under a fixed training budget, as compared to stacked training. It also provided a small but consistent improvement in classification error in the regime of mid-sized training sets. Comparing stacked and gradual training can be viewed as the two extreme adaptation schemes: with stacked-learning reflecting a zero learning rate for the lower layer, and gradual training reflecting a full learning rate. It remains to test intermediate training schedules where the learning rate is being gradually reduced as a layer is presented with examples.

\bibliography{iclr2015}
\bibliographystyle{iclr2015}

\end{document}